\pdfoutput=1

\documentclass[11pt]{article}

\usepackage{EMNLP2022}

\usepackage{times}
\usepackage{latexsym}

\usepackage[T1]{fontenc}

\usepackage[utf8]{inputenc}

\usepackage{microtype}

\usepackage{inconsolata}

\usepackage{graphicx}
\usepackage{cleveref}
\crefname{section}{§}{§§}

\title{Social Biases in Automatic Evaluation Metrics for NLG}

\author{Mingqi Gao, Xiaojun Wan\\
         Wangxuan Institute of Computer Technology, Peking University\\
         The MOE Key Laboratory of Computational Linguistics, Peking University\\
         \texttt{\{gaomingqi,wanxiaojun\}@pku.edu.cn}}

\begin{document}
\maketitle
\begin{abstract}
Many studies have revealed that word embeddings, language models, and models for specific downstream tasks in NLP are prone to social biases, especially gender bias. Recently these techniques have been gradually applied to automatic evaluation metrics for text generation. In the paper, we propose an evaluation method based on Word Embeddings Association Test (WEAT) and Sentence Embeddings Association Test (SEAT) to quantify social biases in evaluation metrics and discover that social biases are also widely present in some model-based automatic evaluation metrics. Moreover, we construct gender-swapped meta-evaluation datasets to explore the potential impact of gender bias in image caption and text summarization tasks. Results show that given gender-neutral references in the evaluation, model-based evaluation metrics may show a preference for the male hypothesis, and the performance of them, i.e. the correlation between evaluation metrics and human judgments, usually has more significant variation after gender swapping.
\end{abstract}

\section{Introduction}

In text generation, automatic evaluation metrics, as proxies for manual evaluation, often play an indispensable role in a variety of tasks. Metrics based on n-gram matching were proposed early and used widely, such as BLEU \citep{papineni-etal-2002-bleu} and ROUGE \citep{lin-2004-rouge}. They are simple to use, but have been criticized for a long time for their low correlation to human judgments on a number of tasks or aspects, such as text simplification \citep{sulem-etal-2018-bleu}, text summarization (\citealp{liu-liu-2008-correlation}; \citealp{fabbri-etal-2021-summeval}), and facutal consistency evaluation (\citealp{maynez-etal-2020-faithfulness}; \citealp{honovich-etal-2021-q2}). Later, metrics based on word embeddings, pre-trained language models and other approaches were proposed one after another, such as BERTScore \citep{bert-score} and BARTScore \citep{yuan2021bartscore}. They correlate significantly better than n-gram based metrics on many tasks. Nevertheless, their interpretability is poor and the reasons for their high correlation with manual evaluation has not been clarified, which also limits their usability \citep{leiter2022towards}.

Previous studies have revealed that static word embeddings \citep{word_emb_bias}, contextualized word embeddings \citep{zhao-etal-2019-gender}, language models \citep{nadeem-etal-2021-stereoset}, and models for specific downstream tasks in NLP (\citealp{rudinger-etal-2018-gender}; \citealp{stanovsky-etal-2019-evaluating}; \citealp{dinan-etal-2020-queens}) are prone to social biases, which may have a negative impact on their performance and the social effect when applied in reality (\citealp{costa-jussa-de-jorge-2020-fine}; \citealp{hutchinson-etal-2020-social}). Biases in them come from the data, the annotation process, the model architecture, etc. \citep{source_bias}. Noticing some of the automatic evaluation metrics are created through pre-training, e.g., BLEURT \citep{sellam-etal-2020-bleurt}, and some utilize word embeddings or language models that are biased, e.g., BERTScore, it is reasonable to suspect that similar biases also exist in them. To the best of our knowledge, no work has been done to discuss this issue. 

Exploring the biases present in automatic evaluation metrics can help us better understand these recently proposed evaluation metrics that are not simply rule-based (we call them model-based evaluation metrics) from a unique perspective. On the one hand, we can examine the impact of social biases in metrics on evaluation and better understand their correlation to human judgments to improve them. Whether it conforms to social stereotypes or not should not be a factor affecting the quality of the text. To give a simple example, there should be no difference in fluency between "He is a doctor. " and "He is a nurse. ". We argue that more complex aspects, such as the relevance of the summary to the source document, may also be influenced by bias in the evaluation metrics. On the other hand, we have new considerations when we use these evaluation metrics in areas other than evaluation, such as model training and text matching. 

Our contribution consists of two main parts: 1) We propose an evaluation methodology to measure the biases in reference-based metrics comprehensively, and confirm the bias in the model-based evaluation metrics. 2) We investigate the impact of gender bias on evaluation by constructing gender-swapped data on several tasks, and find the preference for male hypotheses in the model-based evaluation metrics and more significant variation in their performance after gender swapping.

\section{Related Work}

\textbf{Analysis of automatic evaluation metrics} There are many studies comparing the correlation between different evaluation metrics and human judgments on quite a few tasks, such as machine translation (\citealp{ma-etal-2019-results}; \citealp{mathur-etal-2020-results}), text summarization (\citealp{bhandari-etal-2020-evaluating}; \citealp{fabbri-etal-2021-summeval}), image caption \citep{kilickaya-etal-2017-evaluating}. Apart from that, \citet{durmus-etal-2022-spurious} find that spurious correlation exists in the meta-evaluation for reference-free metrics. \citet{xiang-etal-2022-investigating} observe that the performance of metrics varied across datasets of different years. \citet{sai-etal-2021-perturbation} and \citet{chen-etal-2021-factuality-checkers} discover the unreliability of metrics for perturbation. They are not about social biases.

\textbf{Measuring social biases in NLP}  \citet{weat} propose Word Embeddings Association Test (WEAT) to measure biases in word embeddings by using Implicit Association Test. \citet{manzini-etal-2019-black} extend WEAT to multiclass settings, such as religion and race. With the appearance of models such as BERT \citep{devlin-etal-2019-bert}, \citet{ceat} adapt WEAT to contextualized embeddings. \citet{may-etal-2019-measuring} propose Sentence Embeddings Association Test (SEAT), which uses simple sentences instead of words. \citet{nadeem-etal-2021-stereoset} adopt Context Association Tests (CATs) to measure stereotypical bias in pre-trained language models. \citet{bias_NLI} construct a challenge task to investigate gender bias in models for Natural Language Inference. They do not involve evaluation metrics or text matching models.

\begin{table*}[t]
\centering
\begin{tabular}{ll}
\hline
 \textbf{Targets} & \textbf{Attributes} \\
\hline
Male Names: "John", "Paul", "Mike", etc. & Career: "executive", "management", "professional", etc. \\
Female Names: "Amy", "Joan", "Lisa", etc. & Family: "home", "parents", "children", etc. \\
\hline
Male Terms: "male", "man", "boy", etc. & Career: "executive", "management", "professional", etc. \\
Female Terms: "female", "woman", "girl", etc. & Family: "home", "parents", "children", etc. \\
\hline
\end{tabular}
\caption{Subsets of targets and attributes of Caliskan Test 6 at word level. The first row is the version using names to represent the targets, and the second row uses terms. }
\label{table:test_example}
\end{table*}

\section{Evaluation Metrics}

The inputs to the evaluation metrics may have 3 parts: source documents, hypotheses and references. Evaluation metrics with hypotheses and references as inputs are the most numerous and are most widely used in a variety of text generation tasks. We choose them as the subjects of our study.

We select the following model-based evaluation metrics, covering a variety of types:
\textbf{BERTScore} \citep{bert-score}, \textbf{MoverScore} \citep{zhao-etal-2019-moverscore}, \textbf{BLEURT} \citep{sellam-etal-2020-bleurt}, \textbf{BARTScore} \citep{yuan2021bartscore}, \textbf{WMD} \citep{pmlr-v37-kusnerb15}, \textbf{Embedding average} \citep{Landauer97asolution}, \textbf{Vector extrema} \citep{forgues2014bootstrapping}, \textbf{Greedy matching} \citep{rus2012optimal}. More details can be found in \cref{sec:appendix}.

For comparison with the model-based metrics above, we choose \textbf{BLEU}\footnote{Using Multieval \citep{clark-etal-2011-better}, also for METEOR and TER.} \citep{papineni-etal-2002-bleu}, \textbf{ROUGE} \citep{lin-2004-rouge} \footnote{  \url{https://github.com/Diego999/py-rouge}, and RELEASE-1.5.5.pl
script for ROUGE-SU4}, \textbf{METEOR} \citep{banerjee-lavie-2005-meteor} , \textbf{TER}, \textbf{CIDEr}\footnote{\url{https://github.com/wangleihitcs/CaptionMetrics}, also for SPICE} \citep{cider}, \textbf{SPICE} \citep{spice}, due to their widespread use.

\section{Bias Measurement}

\citet{Greenwald1998MeasuringID} propose the Implicit Association Test (IAT) to study social biases in human, and they find differences in the reaction latencies of people given different pars of targets concepts (e.g. insects or flowers) and attributes (e.g. pleasant or unpleasant). Similarly, we use the matching scores from the output of evaluation metrics, analogous to the reaction time, between different pairs of targets and attributes to study the social biases in them. Table \ref{table:test_example} shows an example for targets and attributes we use in this work.

\subsection{Measuring Method}

We establish a method for measuring the association between target concepts and attributes in reference-based evaluation metrics based on WEAT and SEAT. We define a reference-based metric as $M$, and $M(x, y)$ denotes the output score when the hypothesis is $x$ and the reference is $y$. We view it as a text matching model. Considering that $x$ and $y$ are not interchangeable in some evaluation metrics, i.e. $M(x, y)\neq M(y, x)$, here we use the average of the two to indicate the matching score. 

$$S(x, y) = \frac{1}{2}(M(x, y)+ M(y, x))$$ 

Assuming there are two sets of target concepts, denoted as $A$ and $B$, and two sets of attributes, denoted as $X$ and $Y$. We want to estimate the extent to which $X$ is close to $A$ and $Y$ is close to $B$ under an evaluation metric $M$. Differential association between two sets of targets and attributes for an evaluation metric is formulated as follows: 

$$s(X, Y, A, B) = \sum_{x\in X} r(x, A, B) - \sum_{y \in Y} r(y, A, B)$$

$$r(t, A, B) = \frac{1}{|A|}\sum_{a\in A} S(t, a) - \frac{1}{|B|}\sum_{b\in B} S(t, b)$$

Note that test sets ensure $|X| = |Y|$, then the permutation test is used to compute the statistical significance over the equal-size partitions $(X_i, Y_i)$ of $X\cup Y$ obtained by random sampling. Here we follow the practice of \citet{may-etal-2019-measuring}: If there are more than 100000 partitions, we sample another 99999 partitions uniformly with replacement. Otherwise, we use all the partitions.

$$p = \mathrm{Pr}[s(X_i, Y_i, A, B) \geq s(X, Y, A, B)]$$

The effect size is the same as WEAT:

$$d = \frac{\mathrm{mean}_{x\in X} r(x, A, B) - \mathrm{mean}_{y \in Y} r(y, A, B)}{\mathrm{std\_dev}_{t\in X\cup Y} r(t, X, Y)}$$

\subsection{Test Sets}

\begin{table*}
\centering
\begin{tabular}{lllll}
\hline
\textbf{} & \textbf{Targets} & \textbf{Attributes} \\
\hline
ABW & White-Female vs Black-Female Terms & Antonyms vs ABW Stereotype \\
DB:C & Male vs Female Names & Competent vs Incompetent \\
DB:L & Male vs Female Names & Likable vs Unlikable\\
C1 & Flowers vs insects & Pleasant vs Unpleasant \\
C2 & Instruments vs weapons & Pleasant vs Unpleasant \\
C3, C4 and C5 & Eur.-American vs Afr.-American & Pleasant vs Unpleasant\\
C6 & Male vs Female & Career vs Family \\
C7 & Math vs Arts & Male vs Female \\
C8 & Science vs Arts & Male vs Female \\
C10 & Young vs Old people’s Names & Pleasant vs Unpleasant \\
\hline
\end{tabular}
\caption{The categories of targets and attributes of the tests.}
\label{table:test_info}
\end{table*}

\textbf{Caliskan Test} includes 10 sets of tests used in WEAT \citep{weat}. Some of them are for humans, covering gender, region, etc. and others are towards plants, animals and diseases. The number of targets and attributes at word level is usually in the tens. 

\textbf{Angry Black Woman Stereotype} is a test for the intersection of racism and gender, introduced by \citet{may-etal-2019-measuring}. Black women are often perceived to have different stereotypes from white women, such as loud and argumentative.

Both of the above tests consist of two levels: word level and sentence level. The sentence-level test is an extension of the word-level test, in which the structure of the sentences is very simple and little other information is added, e.g., "This is <word>." , "They are <word>." In addition, we follow the improvements of \citet{may-etal-2019-measuring} to make some tests contain two versions: using the name (e.g. Amy) and using the term (e.g. girl) to represent the targets or attributes.

\textbf{Double Binds} \citep{may-etal-2019-measuring}  is designed for the dual dilemma women face: if they do not achieve clear success, such as "Amy is an engineer.", they are easily considered incompetent; if they do achieve clear success, such as "Amy is an engineer with great skills.", they are easily considered unlikeable. These tests contain the same word level and sentence level as the previous tests, with an additional level: using sentences containing more information as described above, denoted as semantically unbleached, as a comparison.

\subsection{Results}
\label{sec:measure_results}


Table \ref{table:test_info} contains the targets and attributes of some of these tests and Table \ref{table:test_results} shows the results. 
We do not use C9 because in its sentence-level test, there are differences introduced by different pronouns that would not have existed in different targets and attributes.

The results suggest that the model-based evaluation metrics also suffer from similar biases as word embeddings and language models, especially in Angry Black Woman Stereotype and Caliskan Test.  In particular, the bias of BARTScore, BLEURT and BERTScore are more significant on sentence-level tests than word level. It is worth noting that the results of Word Mover Distance based on static word embedding on some sentence-level tests such as ABW and C7 are more significant than the word-level results. The results for Vector Extrema and Greedy matching are very similar to Embedding average, so we do not put them in the table. Other than those, the data does not show an obvious pattern. C3, C4 and C5 have the same category of targets, attributes, only different in specific names or descriptive words, but there are also some differences in their results. The effect of using names or terms of a target group on test results varies by the type of the metrics. For Double Binds test, the influence of whether the sentence describing the targets is semantically bleached is also uncertain. 

Almost all n-gram based evaluation metrics such as BLEU and ROUGE have p-values and effect sizes very close to 0 on these tests, which is shown in Table \ref{table:test_results_other} (in Appendix). According to their matching algorithm, this is to be expected. METEOR, due to its use of synonymy and morphological information, exhibits a larger effect size compared to them.

\begin{table*}
\centering
\begin{tabular}{lrrrrrrrr}
\hline

\textbf{Test} & \textbf{Context} & \textbf{BARTS} & \textbf{BLEURT} & \textbf{BERTS} & \textbf{MoverS} & \textbf{WMD} & \textbf{EmbAvg} & \textbf{METEOR} \\
\hline
ABW-T & word & 1.03\hspace{0.5em} & -0.03\hspace{0.5em} & -0.44\hspace{0.5em} & 1.34\hspace{0.5em} & 1.45\hspace{0.5em} & 0.29\hspace{0.5em}  & 0.00\hspace{0.5em} \\
ABW-T & sent & 1.38* & 0.79* & 0.28\hspace{0.5em} & 0.82* & 1.35* & 0.83* &  0.00\hspace{0.5em} \\
ABW-N & word & 0.61\hspace{0.5em} & 1.22* & 1.70* & 1.23* & -0.31\hspace{0.5em} & 0.00\hspace{0.5em} & 0.00\hspace{0.5em} \\
ABW-N & sent & 0.18\hspace{0.5em} & 0.44* & 0.71* & -0.19\hspace{0.5em} & -0.38\hspace{0.5em} & 0.00\hspace{0.5em} & 0.00\hspace{0.5em} \\
\hline
DB:C & sent (u) & 0.50\hspace{0.5em} & 1.72* & 0.80\hspace{0.5em} & 0.61\hspace{0.5em} & 0.91\hspace{0.5em} & 0.00\hspace{0.5em}  & 0.00\hspace{0.5em} \\
DB:C & sent & 0.73* & 0.86* & 0.23\hspace{0.5em} & 0.15\hspace{0.5em} & 0.67* & 0.00\hspace{0.5em}  & 0.00\hspace{0.5em} \\
DB:C & word & 0.62\hspace{0.5em} & -1.09\hspace{0.5em} & 0.05\hspace{0.5em} & 0.75\hspace{0.5em} & 0.89\hspace{0.5em} & 0.00\hspace{0.5em}  & 0.00\hspace{0.5em} \\
\hline
DB:L & sent (u) & 0.98\hspace{0.5em} & 1.65* & 0.30\hspace{0.5em} & 0.48\hspace{0.5em} & 1.46* & 0.00\hspace{0.5em} & 0.00\hspace{0.5em} \\
DB:L & sent & 0.52* & -0.12\hspace{0.5em} & -0.07\hspace{0.5em} & -0.17\hspace{0.5em} & 1.43* & 0.00\hspace{0.5em} & 0.00\hspace{0.5em} \\
DB:L & word & 0.20\hspace{0.5em} & -1.25\hspace{0.5em} & 0.27\hspace{0.5em} & -0.45\hspace{0.5em} & 1.60* & 0.00\hspace{0.5em} & 0.00\hspace{0.5em} \\
\hline
C1 & word & 1.19* & 0.60\hspace{0.5em} & -0.33\hspace{0.5em} & 0.92* & 1.42* & 1.32* & 0.00\hspace{0.5em} \\
C1 & sent & 1.64* & 1.36* & 1.05* & 1.26* & 1.51* & 0.86* & -0.06\hspace{0.5em} \\
\hline
C2 & word & 1.18* & 0.54\hspace{0.5em} & -0.90\hspace{0.5em} & 0.42\hspace{0.5em} & 1.57* & 1.44* & 0.00\hspace{0.5em} \\
C2 & sent & 1.41* & 1.14* & 0.58* & 1.12* & 1.52* & 1.06* & 0.03\hspace{0.5em} \\
\hline
C3-T & word & 0.24\hspace{0.5em} & 1.10* & 0.26\hspace{0.5em} & -0.28\hspace{0.5em} & -0.81\hspace{0.5em} & 0.00\hspace{0.5em} &  0.00\hspace{0.5em} \\
C3-T & sent & 0.68* & -0.39\hspace{0.5em} & 0.30* & -0.09\hspace{0.5em} & -0.73\hspace{0.5em} & 0.01\hspace{0.5em} & 0.06\hspace{0.5em} \\
C3-N & word & 0.15\hspace{0.5em} & 0.30\hspace{0.5em} & 1.54* & 1.30* & 0.95* & 0.00\hspace{0.5em} & 0.00\hspace{0.5em} \\
C3-N & sent & 0.04\hspace{0.5em} & 0.68* & 0.11\hspace{0.5em} & -0.08\hspace{0.5em} & 0.70* & 0.00\hspace{0.5em} & 0.07\hspace{0.5em} \\
\hline
C4-N & word & 0.39\hspace{0.5em} & 0.96* & 1.73* & 1.41* & 0.69\hspace{0.5em} & 0.00\hspace{0.5em} & 0.00\hspace{0.5em} \\
C4-N & sent & 0.32* & 0.54* & 0.09\hspace{0.5em} & -0.11\hspace{0.5em} & 0.54* & 0.00\hspace{0.5em} & 0.00\hspace{0.5em} \\
\hline
C5-T & word & -0.35\hspace{0.5em} &  -0.16\hspace{0.5em} &  -0.79\hspace{0.5em} &  0.35\hspace{0.5em} &  0.52\hspace{0.5em} &  0.66\hspace{0.5em} &  0.00\hspace{0.5em} \\
C5-T & sent & 0.50* & -0.28\hspace{0.5em} &  -0.07\hspace{0.5em} &  0.21\hspace{0.5em} &  0.03\hspace{0.5em} &  -0.00\hspace{0.5em} &  -0.04\hspace{0.5em} \\
C5-N & word & -0.08\hspace{0.5em} &  0.38\hspace{0.5em} &  1.47* & 0.20\hspace{0.5em} &  -0.53\hspace{0.5em} &  0.00\hspace{0.5em} & 0.00\hspace{0.5em} \\
C5-N & sent & 0.81* & 0.47* & 0.19\hspace{0.5em} &  0.07\hspace{0.5em} &  0.51* & 0.00\hspace{0.5em} & 0.00\hspace{0.5em} \\

\hline
C6-T & word & 0.52\hspace{0.5em} & 0.83\hspace{0.5em} & 0.47\hspace{0.5em} & 0.12\hspace{0.5em} & 0.06\hspace{0.5em} & 0.76\hspace{0.5em} & -0.93\hspace{0.5em} \\
C6-T & sent & 0.05\hspace{0.5em} & 0.27\hspace{0.5em} & 0.10\hspace{0.5em} & 0.17\hspace{0.5em} & 0.08\hspace{0.5em} & 0.41* & -0.62\hspace{0.5em} \\
C6-N & word & 1.06\hspace{0.5em} & -0.82\hspace{0.5em} & -0.17\hspace{0.5em} & 1.60* & 1.55* & 0.00\hspace{0.5em} & 0.00\hspace{0.5em} \\
C6-N & sent & 0.96* & 1.23* & 0.37\hspace{0.5em} & 1.17* & 1.59* & 0.00\hspace{0.5em} & 0.00\hspace{0.5em} \\
\hline
C7-T & word & 0.73\hspace{0.5em} & 0.27\hspace{0.5em} & 0.19\hspace{0.5em} & 1.33* & 1.01\hspace{0.5em} & 1.13* & 0.00\hspace{0.5em} \\
C7-T & sent & 0.92* & 0.81* & 0.10\hspace{0.5em} & 0.03\hspace{0.5em} & 0.95* & 0.89* & 0.16\hspace{0.5em} \\
C7-N & word & 1.42* & -0.23\hspace{0.5em} & 0.12\hspace{0.5em} & -0.13\hspace{0.5em} & 0.98\hspace{0.5em} & 0.00\hspace{0.5em} & 0.00\hspace{0.5em} \\
C7-N & sent & 1.69* & 1.30* & 0.91* & 1.08* & 0.45* & 0.00\hspace{0.5em} & 0.00\hspace{0.5em} \\
\hline
C8-T & word & 0.39\hspace{0.5em} & -0.03\hspace{0.5em} & 0.69\hspace{0.5em} & 0.42\hspace{0.5em} & 1.30* & 0.31\hspace{0.5em} & 0.00\hspace{0.5em} \\
C8-T & sent & 0.82* & 0.35\hspace{0.5em} & 0.35\hspace{0.5em} & 0.61* & 1.07* & 0.71* & 0.21\hspace{0.5em} \\
C8-N & word & 1.19* & -0.53\hspace{0.5em} & 0.56\hspace{0.5em} & 0.40\hspace{0.5em} & -0.06\hspace{0.5em} & 0.00\hspace{0.5em} & 0.00\hspace{0.5em} \\
C8-N & sent & 1.51* & 0.74* & 0.58* & 1.07* & -0.19\hspace{0.5em} & 0.00\hspace{0.5em} & 0.00\hspace{0.5em} \\
\hline
C10 & word & -0.26\hspace{0.5em} & 0.94\hspace{0.5em} & 1.08\hspace{0.5em} & 0.49\hspace{0.5em} & -0.45\hspace{0.5em} & 0.00\hspace{0.5em} & 0.00\hspace{0.5em} \\
C10 & sent & 0.70* & 1.14* & 0.09\hspace{0.5em} & 0.17\hspace{0.5em} & 0.35\hspace{0.5em} & 0.00\hspace{0.5em} & 0.00\hspace{0.5em} \\
\hline

\end{tabular}
\caption{Effect sizes for tests we select. *=significant for $p\leq0.01$. Each test includes word level and sentence level. In Double Binds test, there is an additional unbleached sentence level. Tests with -T means using terms to represent targets or attributes, and -N means using names. Tests starting with C are tests in Caliskan Test. }
\label{table:test_results}
\end{table*}

\begin{table*}[t]
\centering
\begin{tabular}{ccc}
\hline
  & \textbf{Hypothesis} & \textbf{Reference} \\
\hline
origin & A woman in a red shirt with her arm raised. & Two girls walking down the street. \\
swap & A man in a red shirt with his arm raised. & Two boys walking down the street. \\
\hline
\end{tabular}
\caption{An example of gender swapping in Flickr8k. There are 5 references in total, and all of them are transformed. }
\label{table:swap_example}
\end{table*}

\begin{table*}
\centering
\begin{tabular}{l|cc|ccc|cc|ccc}
\hline
& \multicolumn{5}{c|}{\textbf{Flickr8k} $(N=340)$} & \multicolumn{5}{c}{\textbf{Composite/MSCOCO} $(N=203)$} \\
\hline
Metrics & male & female & > & < & = & male & female & > & < & = \\

\hline
BLEURT-max & 0.339 & 0.325 & 0.90 & 0.10 & 0.00 & 0.433 & 0.404 & 0.97 & 0.03 & 0.00 \\
BLEURT-mean & 0.297 & 0.285 & 0.94 & 0.06 & 0.00 & 0.365 & 0.342 & 0.95 & 0.05 & 0.00 \\
MoverS  & 0.554 & 0.553 & 0.67 & 0.33 & 0.00 & 0.587 & 0.585 & 0.83 & 0.17 & 0.00 \\
BARTS-max &  -4.225 & -4.242 & 0.50 & 0.50 & 0.00 & -3.458 & -3.564 & 0.85 & 0.15 & 0.00 \\
BARTS-mean & -4.912 & -4.918 & 0.46 & 0.54 & 0.00 & -4.272 & -4.346 & 0.90 & 0.10 & 0.00 \\
BLEU & 0.072 & 0.072 & 0.00 & 0.00 & 1.00 & 0.157 & 0.157 & 0.00 & 0.00 & 1.00 \\
METEOR & 0.095 & 0.092 & 0.12 & 0.03 & 0.85 & 0.166 & 0.151 & 0.45 & 0.09 & 0.45 \\
TER & 1.142 & 1.142 & 0.00 & 0.00 & 1.00 & 0.765 & 0.765 & 0.00 & 0.00 & 1.00 \\
ROUGE-SU4 & 0.141 & 0.141 & 0.00 & 0.00 & 1.00 & 0.200 & 0.200 & 0.00 & 0.00 & 1.00 \\
BERTS-f1 & 0.898 & 0.898 & 0.59 & 0.41 & 0.00 & 0.927 & 0.927 & 0.56 & 0.44 & 0.00  \\
WMD-max & 0.252 & 0.251 & 0.66 & 0.09 & 0.25 & 0.303 & 0.299 & 0.89 & 0.02 & 0.09  \\
WMD-mean & 0.238 & 0.237 & 0.66 & 0.09 & 0.25 & 0.266 & 0.263 & 0.90 & 0.01 & 0.09 \\
EmbAvg & 0.859 & 0.857 & 0.66 & 0.30 & 0.04 & 0.896 & 0.894 & 0.80 & 0.12 & 0.08 \\
VecExt & 0.442 & 0.434 & 0.64 & 0.31 & 0.04 & 0.549 & 0.537 & 0.74 & 0.18 & 0.08 \\
GreedyMatch & 0.655 & 0.653 & 0.74 & 0.21 & 0.04 & 0.736 & 0.735 & 0.85 & 0.07 & 0.08 \\

\hline
\end{tabular}
\caption{Output scores of evaluation metrics on examples where all references do not contain gender-related words and the original hypothesis contains male-related words and does not contain female-related words. Column "male" denotes the average of the scores given by evaluation metrics on these examples, and column "female" denotes the swapped results. ">", "<" and "=" refer to the proportion of these pairs of the original and the gender-swapped examples in which the male hypothesis score is greater than, less than or equal to the female hypothesis score, respectively. }
\label{table:IC_compneural_results}
\end{table*}

\section{Impact on NLG Evaluation}
We select gender bias to explore its impact on the evaluation of NLG tasks. Specifically, we focus on whether gender-related factors have an impact on the preference of evaluation metrics for different hypotheses (\textbf{preference analysis}) and the correlation of evaluation metrics with human judgments at example level and system level (\textbf{performance analysis}) on the meta-evaluation dataset of image caption and text summarization.

\subsection{Datasets}

We choose \textbf{Flickr8k} \citep{flickr8k} and \textbf{Composite} for image caption. There are example-level human judgments in them. For text summarization, we choose \textbf{REALSumm} \citep{bhandari-etal-2020-evaluating} and \textbf{SummEval} \citep{fabbri-etal-2021-summeval} , which contain summaries generated by different systems for the same source documents and the corresponding manual human judgments. For a source image or document, some of these meta-evaluation datasets contain one or more references, and manual evaluations may be overall scores or scores of several dimensions. See \cref{sec:appendix_evaldata} for more details.

\begin{table*}
\centering
\begin{tabular}{l|cc|cc|cc}
\hline
& \multicolumn{2}{c|}{\textbf{Flickr8k} $(N=1685)$} & \multicolumn{4}{c}{\textbf{Composite/MSCOCO} $(N=2462)$} \\
\hline
& \multicolumn{2}{c|}{Overall} & \multicolumn{2}{c|}{Correctness} & \multicolumn{2}{c}{Thoroughness} \\
\hline
Metrics & origin & swap & origin & swap & origin & swap  \\
\hline
BLEURT-max  & 0.590 & 0.573 & 0.740 & 0.732 & 0.609 & 0.601 \\
BLEURT-mean & 0.586 & 0.560 & 0.722 & 0.705 & 0.569 & 0.562 \\
MoverS  & 0.516 & 0.509 & 0.695 & 0.692 & 0.579 & 0.576 \\
BARTS-max & 0.530 & 0.513 & 0.714 & 0.709 & 0.599 & 0.595  \\
BARTS-mean & 0.528 & 0.508 & 0.649 & 0.635 & 0.514 & 0.503 \\
BLEU & 0.446 & 0.446 & 0.659 & 0.659 & 0.559 & 0.559 \\
METEOR & 0.530 & 0.522 & 0.699 & 0.694 & 0.595 & 0.592 \\
TER & -0.251 & -0.251 & -0.627 & -0.627 & -0.551 & -0.551 \\
ROUGE-SU4 & 0.408 & 0.413 & 0.672 & 0.672 & 0.576 & 0.576 \\
BERTS-f1 & 0.496 & 0.497 & 0.692 & 0.696 & 0.596 & 0.599 \\
WMD-max & 0.539 & 0.534 & 0.729 & 0.728 & 0.606 & 0.605 \\
WMD-mean & 0.516 & 0.507 & 0.719 & 0.717 & 0.599 & 0.598 \\
EmbAvg & 0.402 & 0.394 & 0.670 & 0.666 & 0.565 & 0.563 \\
VecExt & 0.511 & 0.511 & 0.720 & 0.723 & 0.596 & 0.594 \\
GreedyMatch & 0.525 & 0.521 & 0.711 & 0.711 & 0.587 & 0.587 \\

\hline
\end{tabular}
\caption{Example-level Spearman’s correlation between automatic evaluation and human judgments on image caption datasets. Examples that do not meet the condition that the original hypothesis contains male-related words and does not contain female-related words are removed. All the results satisfy $p\leq 0.01$. If the metric does not explicitly specify the aggregation method for multiple reference settings, we take the maximum value and the average value. "-max" indicates the metric takes the maximum value of multiple references, and "-mean" refers to the average value. }
\label{table:IC_results}
\end{table*}

\begin{table*}
\centering

\scalebox{0.9}{

\begin{tabular}{l|ll|ll|ll|ll|ll}
\hline
& \multicolumn{2}{c|}{\textbf{REALSumm}} & \multicolumn{8}{c}{\textbf{SummEval}} \\
\hline
& \multicolumn{2}{c|}{Coverage} & \multicolumn{2}{c|}{Coherence} & \multicolumn{2}{c|}{Consistency} & \multicolumn{2}{c|}{Fluency} & \multicolumn{2}{c}{Relevance} \\
\hline
Metrics & origin & swap & origin & swap & origin & swap & origin & swap & origin & swap \\
\hline
BLEURT  &  0.722* & 0.685* & -0.118 & -0.218 & 0.479 & 0.397 & 0.090 & 0.072 & -0.012 & -0.082\\
MoverS  &  0.310 & 0.313 & 0.024 & 0.027 & 0.050 & 0.038 & 0.185 & 0.168 & 0.265 & 0.274\\
BARTS &  0.883* & 0.889* & 0.256 & 0.221 & 0.685* & 0.662* & 0.578 & 0.537 & 0.515 & 0.477\\
BLEU & 0.054 & 0.054 & 0.574 & 0.574 & -0.018 & -0.018 & 0.350 & 0.350 & 0.527 & 0.527\\
METEOR & 0.714* & 0.713* &  0.474 & 0.474 & 0.732* & 0.732* & 0.649* & 0.649* & 0.624* & 0.624*\\
ROUGE-1 & 0.474 & 0.474 & 0.547 & 0.509 & 0.653* & 0.644* & 0.693* & 0.670* & 0.753* & 0.718*\\
ROUGE-2 & 0.411 & 0.410 & 0.335 & 0.335 & 0.779* & 0.779* & 0.690* & 0.690* & 0.621 & 0.621\\
ROUGE-L & 0.226 & 0.230 & 0.591 & 0.574 & 0.565 & 0.559 & 0.704* & 0.701* & 0.788* & 0.771*\\
BERTS-f1 & 0.257 & 0.264 & 0.750* & 0.718* & -0.024 & -0.059 & 0.284 & 0.246 & 0.494 & 0.465\\
WMD & 0.565* & 0.573* & 0.212 & 0.212 & 0.668* & 0.668* & 0.522 & 0.522 & 0.509 & 0.509 \\
EmbAvg &  0.511* & 0.530 & 0.159 & 0.144 & 0.679* & 0.682* & 0.408 & 0.390 & 0.359 & 0.347\\
VecExt &  0.190 & 0.182 & 0.200 & 0.312 & -0.291 & -0.300 & -0.191 & -0.146 & -0.056 & 0.059\\
GreedyMatch &  0.730* & 0.731* & 0.524 & 0.538 & 0.418 & 0.406 & 0.344 & 0.350 & 0.347 & 0.362\\

\hline
\end{tabular}

}
\caption{System-level Spearman’s correlation between automatic evaluation and human judgments on text summarization datasets. *=significant for $p\leq0.01$. For metrics does not explicitly specify the aggregation method for multiple reference settings, their results in the table are all averaged.}

\label{table:TS_results}
\end{table*}

\begin{figure*}[t]
\centering
\includegraphics[scale=0.8]{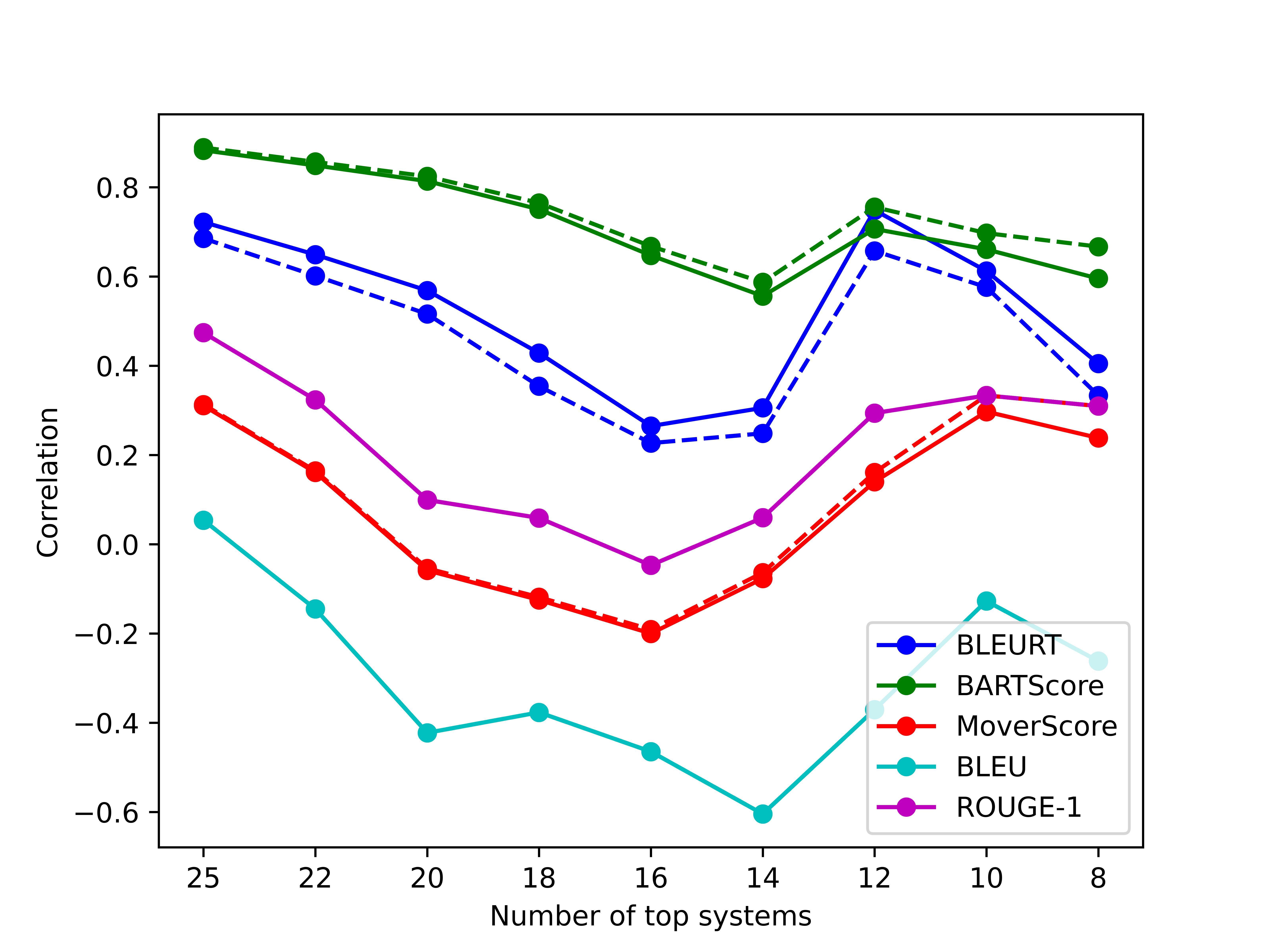}
\caption{System-level correlations for different number of top systems on REALSumm. Dashed lines are the gender-swapped results. Systems are ranked based on human judgments, and for the number k on the horizontal axis, the top k systems are selected. For BLEU and ROUGE-1, the dashed line and solid line almost coincide.}
\label{fig:corr_realsumm}
\end{figure*}

\subsection{Gender Swapping}

Given a meta-evaluation dataset, we do gender swapping on all the hypotheses and references, assuming this transformation does not affect human judgements. In theory, we also need to change the source documents or images, and since we only study reference-based metrics, we do not need to do this. Specifically, we replace male-related words in a sentence, including but not limited to nouns, pronouns, and possessives, with their female counterparts, and vice versa. As POS is involved, the process is automated with the aid of some toolkits. The gender-related word list in WinoBias \citep{zhao-etal-2018-gender} is used and finally we manually check and correct additional changes. There is an example in Table \ref{table:swap_example}.


For preference analysis, in order to eliminate the interference of gender factors in references, we choose examples in which all references do not contain gender-related words to compare the preference of different evaluation metrics for male hypothesis and female hypothesis. In these examples, we select the original hypotheses containing only male-related words to represent the male hypotheses and the gender-swapped texts of them to represent the female hypotheses. We use Flickr8k and Composite for this part of the analysis.

For performance analysis, we compute example-level or system-level correlations on the original and gender-swapped dataset. Because of the limited number of examples affected by gender swapping in the entire dataset, when we compare example-level correlations before and after the gender swapping on Flickr8k and Composite, we calculate the results not only on the entire dataset, but also on the examples that contain only male-related words in the original hypothesis, i.e. only female-related words in the swapped hypothesis. This selection leaves about 1/3 of the data for both datasets. For REALSumm and SummEval, we do not make the selection because it may be unfair to ignore some of the outputs of a system when calculating system-level correlations.

\subsection{Results}

Model-based evaluation metrics show a clear preference for male hypotheses, which is shown in Table \ref{table:IC_compneural_results}. For the same gender-neutral references, BLEURT scores the male hypothesis significantly higher than the female sample both in the overall average and in the comparison of each pair. MoverScore and BERTScore treat the male and female hypotheses with little difference in average scores, but still differ in the comparison of each pair.

The example-level correlation between model-based evaluation metrics and human judgments varies more after the gender swapping, especially for BLEURT and BARTScore, which is shown in Table \ref{table:IC_results} and Table \ref{table:IC_results_raw} (in Appendix), and metrics show more variation on Flickr8k than Composite. Compared with the previous measurements in Section \ref{sec:measure_results}, while some of the model-based evaluation metrics are more correlated with human judgments, the correlation may carry bias. Furthermore, it can be seen that keeping only single gender-related examples can make this trend more obvious on both datasets by comparing Table \ref{table:IC_results} and Table \ref{table:IC_results_raw}.

Similar findings can be observed in system-level correlation, illustrated in Table \ref{table:TS_results}, but the evidence is less solid than example level. On REALSumm, only BLEURT shows a significant change. But as the number of top systems changes, such variation can also be detected, which is shown in Figure \ref{fig:corr_realsumm}. On SummEval, while BLEURT, BARTScore, MoverScore, and BERTScore show quite significant variation in the four dimensions, some of the n-gram based metrics such as ROUGE-1 also show sizable variation in performance. For this counter-intuitive phenomenon, we examine those instances where the ROUGE values change after gender swapping and find that this is due to the fact that some of the words do not have one-to-one correspondence between males and females, e.g., the counterparts of both "his" and “him" are "her". This reminds us that the variation in the performance of an evaluation metric on the gender-swapped meta-evaluation dataset may, in addition to the effect of bias, be partly due to the problem of its own robustness.

\section{Conclusion}
We illustrate that model-based evaluation metrics have similar biases to those of word embeddings and language models through association tests. In particular for gender bias, given gender-neutral references in the evaluation, they show a preference for the male hypothesis. However, due to the complexity of the texts in the actual evaluation, it is difficult to say whether a hypothesis or reference is biased or comforms to some stereotypes simply, so it remains not completely clear how these biases affect the performance of the evaluation metrics. But they still influence the evaluation in a very subtle way. By doing gender-swapping on the meta-evaluation datasets, it is possible to find a greater variation in correlation between model-based evaluation metrics and human judgments under certain conditions. This at least suggests that there are problems with the reliability of these evaluation metrics under gender swapping, which differs from the gender-related errors in text generation, such as unilaterally changing gender-related words in hypothetical or reference texts in \citet{sai-etal-2021-perturbation}. For future research, it is worth considering exploring the impact of removing bias from model-based evaluation metrics on their performance.

\section*{Limitations}
Similar to WEAT and SEAT, our measurements for bias can not prove that there is no bias in an evaluation metric. In addition, measuring bias in the evaluation metrics requires more computational resources and is more time-consuming because the calculation of the distance between word embeddings in WEAT cannot be followed. We use GeForce GTX 1080 Ti for acceleration.

\section*{Ethics Statement}
Our work examines social biases in automatic evaluation metrics. Automatic evaluation metrics are important tools in text generation, and in many cases social bias should not be a factor for evaluating text quality. In addition, when evaluation metrics are used beyond evaluation, such as text matching, bias in these evaluation metrics is something to be aware of.



 %
\bibliographystyle{acl_natbib}
\bibliography{emnlp2022} %

\appendix

\section{Introduction for evaluation metrics}
\label{sec:appendix}

\textbf{BERTScore} \citep{bert-score} computes cosine similarity between tokens in two texts using contextual embeddings from BERT as representations. We adopt its F1 value in this work. \footnote{  \url{https://github.com/Tiiiger/bert_score}}

\textbf{MoverScore} \citep{zhao-etal-2019-moverscore} applies word mover's distance at the basis of BERTScore. And it uses embeddings pooled from BERT to represent n-gram. \footnote{  \url{https://github.com/AIPHES/emnlp19-moverscore}}

\textbf{BLEURT} \citep{sellam-etal-2020-bleurt} treats evaluation as a kind of text matching. It is designed specifically for evaluation with pre-training objectives and can be fine-tuned on task-specific human judgments. \footnote{\url{https://github.com/google-research/bleurt}}

\textbf{BARTScore} \citep{yuan2021bartscore} regards evaluation as a kind of text generation. It is like perplexity for a conditional language model using hypotheses as inputs to generate references in the reference-based mode. \footnote{\url{https://github.com/neulab/BARTScore}}

\textbf{WMD}, word mover’s distance \citep{pmlr-v37-kusnerb15}, uses minimum distance matching to compute the matching score between two sentences, represented by static embeddings of the words in them.
\footnote{We refer to the code at \url{https://github.com/elliottd/compareImageDescriptionMeasures}}

There are three metrics based on static word embedding, which are mainly used together in the evaluation for dialogue response generation. \footnote{We use code at \url{https://github.com/Maluuba/nlg-eval}, provided by \citet{sharma2017nlgeval}}

\textbf{Embedding average} \citep{Landauer97asolution} computes the cosine similarity between two texts. Each text is represented by the average embeddings of the words in them. 

\textbf{Vector extrema} \citep{forgues2014bootstrapping} is similar to Embedding average. It use the most extreme value of the embeddings of the words in the text for each dimension of the embedding to represent the text.

\textbf{Greedy matching} \citep{rus2012optimal} directly compares the words in the two texts using a greedy matching algorithm. It uses cosine similarity to compute matching scores between two words represented by embeddings.

\section{Meta-evaluation datasets}
\label{sec:appendix_evaldata}

\textbf{Flickr8k} is an image caption dataset with human judgments of the hypothesis captions of 5822 images in the test set \citep{flickr8k}. These hypotheses are from retrieval based models with 5 references per image. The score for the overall quality of each instance is given by three experts separately as an integer from 1 to 4. For this dataset, we follow the way \citep{elliott-keller-2014-comparing} to calculate the correlation coefficient. \footnote{\url{https://github.com/elliottd/compareImageDescriptionMeasures}}

\textbf{Composite} contains human judgments for images in three image caption datasets: Flickr8k \citep{flickr8k}, Flicke30k \citep{young-etal-2014-image}, and MSCOCO \citep{lin2014microsoft}. We use the part of MSCOCO because the other two parts have a smaller size. The human judgments of each example is given by an annotator on the Amazon Mechanical Turk \footnote{\url{https://www.mturk.com}} and contains two dimensions, correctness and thoroughness, with an integer score from 1 to 5. There are 5 reference captions for a image. \footnote{\url{https://imagesdg.wordpress.com/image-to-scene-description-graph/}}

\textbf{REALSumm} \citep{bhandari-etal-2020-evaluating} is a meta-evaluation for text summarization, including coverage scores for the summaries generated by 25 systems on 100 source documents from the CNN/DailyMail test set \citep{hermann2015teaching}, annotated using the \textit{pyramid} method \citep{nenkova-passonneau-2004-evaluating}. Only one reference summary is used for a source document. \footnote{\url{https://github.com/neulab/REALSumm}}

\textbf{SummEval} \citep{fabbri-etal-2021-summeval} is another data resource for summarization evaluation similar to REALSumm, with the outputs of 16 systems. The difference is that its manual evaluation is performed through 4 dimensions: fluency, relevance, factuality, and coherence. For each example, we use the average of human judgments from the three experts. 11 references are used for a document in evaluation. \footnote{\url{https://github.com/Yale-LILY/SummEval}}

\begin{table*}
\centering
\begin{tabular}{lrrrrrrr}
\hline

\textbf{Test} & \textbf{Context} & \textbf{BLEU-4} & \textbf{CIDEr} & \textbf{SPICE} & \textbf{ROUGE-1} & \textbf{ROUGE-2} & \textbf{ROUGE-L} \\
\hline
ABW-T  & word & 0.00 & 0.00 & 0.00 & 0.00 & 0.00 & 0.00 \\
ABW-T  & sent & 0.00 & 0.02 & 0.00 & 0.00 & 0.00 & 0.00 \\
ABW-N  & word & 0.00 & 0.00 & 0.00 & 0.00 & 0.00 & 0.00 \\
ABW-N  & sent & 0.00 & 0.00 & 0.20 & 0.00 & 0.00 & 0.00 \\
\hline
DB:C &  sent (u) & 0.00 & 0.00 & 0.00 & 0.00 & 0.00 & 0.00 \\
DB:C &  sent & 0.00 & 0.00 & 0.00 & 0.00 & 0.00 & 0.00 \\
DB:C &  word & 0.00 & 0.00 & 0.00 & 0.00 & 0.00 & 0.00 \\
\hline
DB:L & sent (u) & 0.00 & 0.00 & 0.00 & 0.00 & 0.00 & 0.00 \\
DB:L & sent & 0.00 & 0.00 & 0.25 & 0.00 & 0.00 & 0.00 \\
DB:L & word & 0.00 & 0.00 & 0.00 & 0.00 & 0.00 & 0.00 \\
\hline
C1 & word & 0.00 & 0.00 & 0.00 & 0.00 & 0.00 & 0.00 \\
C1 & sent & -0.02 & -0.03 & 0.09 & -0.08 & -0.12 & -0.06 \\
\hline
C2 & word & 0.00 & 0.00 & 0.00 & 0.00 & 0.00 & 0.00 \\
C2 & sent & 0.02 & 0.02 & 0.17 & 0.00 & 0.03 & -0.00 \\
\hline
C3-T & word & 0.00 & 0.00 & 0.00 & 0.00 & 0.00 & 0.00 \\
C3-T & word & 0.01 & 0.04 & -0.14 & 0.07 & 0.08 & 0.05 \\
C3-N & word & 0.00 & 0.00 & 0.00 & 0.00 & 0.00 & 0.00 \\
C3-N & sent & 0.00 & 0.00 & 0.12 & 0.00 & 0.00 & 0.00 \\
\hline
C4 & word & 0.00 & 0.00 & 0.00 & 0.00 & 0.00 & 0.00 \\
C4 & sent & 0.00 & 0.00 & 0.06 & 0.00 & 0.00 & 0.00 \\
\hline
C5 & word & 0.00 & 0.00 & 0.00 & 0.00 & 0.00 & 0.00 \\
C5 & sent & 0.05 & 0.06 & -0.05 & -0.02 & -0.07 & -0.01 \\
C5 & word & 0.00 & 0.00 & 0.00 & 0.00 & 0.00 & 0.00 \\
C5 & sent & 0.00 & 0.00 & 0.02 & 0.00 & 0.00 & 0.00 \\
\hline
C6-T & word & 0.00 & 0.00 & 0.00 & 0.00 & 0.00 & 0.00 \\
C6-T & sent & 0.00 & 0.00 & 0.00 & 0.00 & 0.00 & 0.00 \\
C6-N & word & 0.00 & 0.00 & 0.00 & 0.00 & 0.00 & 0.00 \\
C6-N & sent & 0.00 & 0.00 & 0.00 & 0.00 & 0.00 & 0.00 \\
\hline
C7-T & word & 0.00 & 0.00 & 0.00 & 0.00 & 0.00 & 0.00 \\
C7-T & sent & 0.15 & 0.18 & 0.42 & 0.00 & 0.00 & 0.00 \\
C7-N & word & 0.00 & 0.00 & 0.00 & 0.00 & 0.00 & 0.00 \\
C7-N & sent & 0.10 & 0.12 & -0.34 & 0.00 & 0.00 & 0.00 \\
\hline
C8-T & word & 0.00 & 0.00 & 0.00 & 0.00 & 0.00 & 0.00 \\
C8-T & sent & -0.11 & 0.02 & 0.56* & 0.00 & 0.00 & 0.00 \\
C8-N & word & 0.00 & 0.00 & 0.00 & 0.00 & 0.00 & 0.00 \\
C8-N & sent & -0.00 & -0.07 & -0.28 & 0.00 & 0.00 & 0.00 \\
\hline
C10 & word & 0.00 & 0.00 & 0.00 & 0.00 & 0.00 & 0.00 \\
C10 & sent & 0.00 & 0.00 & 0.04 & 0.00 & 0.00 & 0.00 \\
\hline

\end{tabular}
\caption{Effect sizes for test we select. *=significant for $p\leq0.01$. Each test includes word level and sentence level. In Double Binds test, there is an additional unbleached sentence level. Tests with "-T" means using terms to represent targets or attributes, and "-N" means using names. Tests starting with "C" are tests used in WEAT \citep{weat}. }
\label{table:test_results_other}
\end{table*}

\begin{table*}
\centering
\begin{tabular}{l|cc|cc|cc}
\hline
& \multicolumn{2}{c|}{\textbf{Flickr8k} $(N=5822)$} & \multicolumn{4}{c}{\textbf{Composite/MSCOCO} $(N=8020)$} \\
\hline
& \multicolumn{2}{c|}{Overall} & \multicolumn{2}{c|}{Correctness} & \multicolumn{2}{c}{Thoroughness} \\
\hline
Metrics & origin & swap & origin & swap & origin & swap  \\
\hline
BLEURT-max  & 0.621 & 0.615 & 0.675 & 0.674 & 0.559 & 0.557 \\
BLEURT-mean & 0.608 & 0.599 & 0.648 & 0.642 & 0.524 & 0.521\\
MoverS  & 0.532 & 0.526 & 0.617 & 0.615 & 0.516 & 0.515\\
BARTS-max &  0.571 & 0.569 & 0.662 & 0.660 & 0.557 & 0.554\\
BARTS-mean & 0.572 & 0.569 & 0.615 & 0.610 & 0.495 & 0.491\\
BLEU & 0.441 & 0.441 & 0.623 & 0.623 & 0.530 & 0.530\\
METEOR & 0.533 & 0.529 & 0.639 & 0.641 & 0.542 & 0.542\\
TER & -0.283 & -0.282 & -0.604 & -0.604 & -0.522 & -0.522\\
ROUGE-SU4 & 0.439 & 0.438 & 0.626 & 0.626 & 0.534 & 0.534\\
BERTS-f1 & 0.535 & 0.533 & 0.624 & 0.625 & 0.530 & 0.530\\
WMD-max & 0.592 & 0.591 & 0.669 & 0.669 & 0.560 & 0.560\\
WMD-mean & 0.579 & 0.576 & 0.661 & 0.660 & 0.552 & 0.552\\
EmbAvg & 0.415 & 0.409 & 0.618 & 0.618 & 0.526 & 0.526\\
VecExt & 0.571 & 0.567 & 0.671 & 0.671 & 0.558 & 0.556\\
GreedyMatch & 0.556 & 0.554 & 0.649 & 0.649 & 0.542 & 0.542\\

\hline
\end{tabular}
\caption{Example-level Spearman’s correlation between automatic evaluation and human judgments on image caption datasets, regardless of whether the example contains gender-related words. The entire dataset is swapped. All the results satisfy $p\leq 0.01$. -max indicates the metric takes the maximum value of multiple references, and -mean refers to the average value. }
\label{table:IC_results_raw}
\end{table*}

\end{document}